\pdfoutput=1
\documentclass[conference,a4paper]{IEEEtran}

\usepackage[table]{xcolor} 
\usepackage[utf8]{inputenc}
\usepackage[T1]{fontenc}
\usepackage{graphicx}
\usepackage{amsmath}
\usepackage{tikz}
\usepackage{todonotes} 
\usepackage{floatrow} 
\usepackage{booktabs}
\usepackage{amssymb}
\usepackage{tabularx} 
\usepackage{array} 
\usepackage{pifont} 
\usepackage{seqsplit}
\usepackage{hyperref} 
\newcommand{\cmark}{\ding{51}} 
\newcommand{\xmark}{\ding{55}}

\usetikzlibrary{positioning}

\title{Assembly-JEPA: Using World Models for Fine-Grained Industrial Image Verification}

\author{
\IEEEauthorblockN{Michael Hollmer}
\IEEEauthorblockA{\textit{Deggendorf Institute of Technology}\\
Deggendorf, Germany\\
\textit{Zollner Elektronik AG}\\
Zandt, Germany\\
michael.hollmer@stud.th-deg.de}
\and
\IEEEauthorblockN{Andreas Fischer}
\IEEEauthorblockA{\textit{Deggendorf Institute of Technology}\\
Deggendorf, Germany\\
andreas.fischer@th-deg.de}
}

\begin{document}
\maketitle

\begin{abstract} 
Automated verification of industrial assembly processes remains a challenging problem due to substantial visual variability, limited availability of labeled defect data, and the fine-grained nature of many assembly errors. Existing approaches often rely on object detection, CAD models, or supervised classification pipelines, which can struggle to generalize across changing products and assembly configurations.

In this work, we propose a novel self-supervised assembly verification framework based on Joint Embedding Predictive Architectures (JEPAs). Starting from the JEPA-based world model LeWorldModel, we investigate how the architecture must be adapted to verify the correctness of fine-grained assembly operations. To this end, we formulate assembly verification as a predictive world modeling task operating on before-and-after observations of assembly processes.

Experimental results on industrial assembly tasks demonstrate that the proposed approach reliably detects both fine-grained incorrect assembly states and action-state mismatches, achieving an AUROC of 86.9\% on a newly introduced assembly verification dataset. These results indicate that JEPA-based predictive representations constitute a promising foundation for robust and data-efficient industrial assembly verification. 

\end{abstract}


\section{Introduction}
Industrial assembly verification remains a challenging problem despite significant advances in computer vision and machine learning. Machine-learning-based vision systems have become an important component of modern manufacturing, supporting quality control, automated optical inspection and process monitoring. Recent surveys report a rapidly growing body of research and industrial applications in deep-learning-based visual inspection \cite{asi7010011}.

While existing methodologies, such as CNN-based classification (ResNet \cite{ResNet}), object detection (YOLO \cite{Yolo}), or anomaly detection (PaDiM \cite{Padim}, PatchCore \cite{PatchCore}), perform remarkably well on standardized benchmarks like MVTec AD \cite{MVTec}, recent evaluation studies emphasize that significant challenges remain for real-world industrial deployment \cite{BeyondAcademicBenchmarks}. Changes in illumination, background clutter, and object appearance can considerably affect system performance. Consequently, high inspection accuracy is often achieved by restricting deployment to controlled environments with stable illumination and background conditions, avoiding unconstrained variations \cite{asi7010011, BeyondAcademicBenchmarks}. 

In the assembly of large and complex products such as electrical cabinets, however, maintaining a fixed camera position and perfectly consistent lighting conditions is often impractical or even impossible. Furthermore, assembly errors are frequently highly localized and fine-grained, making reliable verification particularly challenging for conventional approaches.

Joint Embedding Predictive Architectures (JEPAs) have emerged as an effective paradigm for learning semantically meaningful visual representations \cite{IJepa}. In computer vision, JEPA-based training is primarily used to obtain powerful image encoders whose features can subsequently be applied to downstream tasks such as classification or anomaly detection. In most of these applications, however, only the encoder is retained after training, while the predictor network is discarded and serves merely as a pretraining component.

In contrast, research on world models preserves the complete JEPA architecture and utilizes the predictor to model state transitions in a latent representation space. Given a conditional signal, such as an action to be executed by a robot, the predictor learns to estimate future states and thereby captures the dynamics of the underlying system. This ability has made JEPA-based world models a promising framework for planning, control, and predictive reasoning.

In this paper, we aim to bridge the gap between these two research directions and propose a JEPA-based model for fine-grained industrial assembly verification. Unlike prior JEPA-based computer vision approaches that primarily use the predictor during training, our method retains both the encoder and predictor during inference. By conditioning the predictor on assembly actions, the model learns expected state transitions between consecutive assembly steps. Verification is then performed by comparing the predicted post-action state with the actually observed state, using the prediction error as an indicator of assembly correctness. To the best of our knowledge, this is the first work to leverage the complete JEPA, including both encoder and predictor, for industrial assembly verification. We will publicly release the source code, model checkpoints, and datasets to support reproducibility.

\section{Related Work}
Self-supervised learning approaches based on JEPAs have demonstrated strong performance in visual representation learning. Assran et al. introduced I-JEPA and showed that predictive latent-space objectives can produce semantically meaningful image representations while achieving competitive performance on downstream tasks \cite{IJepa}. Building on this idea, Balestriero and LeCun introduced latent-space regularization techniques that prevent representational collapse without relying on heuristic solutions while further improving downstream performance \cite{balestriero2025lejepaprovablescalableselfsupervised}. More recently, Maes et al. extended these concepts to world modeling and demonstrated that the same regularization principles can be used to learn environment dynamics directly in latent space \cite{maes2026leworldmodelstableendtoendjointembedding}. These approaches, however, focus on general visual representation learning and world modeling rather than the specific task of assembly inspection.

In contrast, existing approaches to automated assembly inspection frequently rely on CAD information, extensive manual annotations, predefined component categories, or combinations thereof. Early assembly inspection approaches in particular were predominantly based on CAD representations of products and assemblies. Sopauschke et al. employ 3D sensing and CAD reference models for automated quality inspection by comparing the observed assembly state against the intended design \cite{AutomatedQualityInspection}. Likewise, Bründl et al. leverage CAD data to semantically segment control cabinet components and identify assembly-relevant geometric features that can support automated assembly processes \cite{SemanticPartSegmentation}.

\begin{figure*}[t] 
\centering 
\includegraphics[width = 400pt]{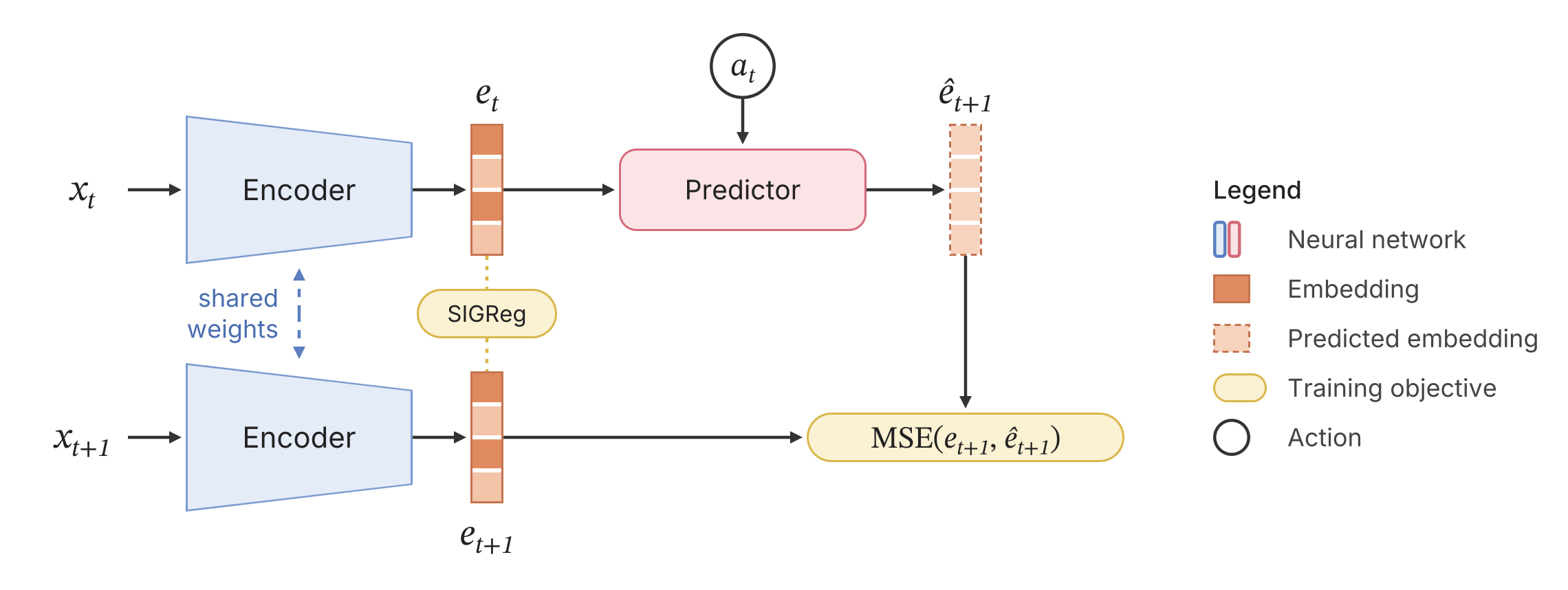} 
\caption{Overview of the LeWorldModel architecture \cite{maes2026leworldmodelstableendtoendjointembedding}, including the encoder, action-conditioned predictor, and SIGReg regularization of the latent representations.}
\label{fig:leworldmodel}
\end{figure*}

In addition to CAD-based approaches, several studies employ deep learning methods operating directly on 2D image data. Scheck et al. utilize object detection techniques to identify and localize components in electrical control cabinets, supporting automated disassembly processes \cite{DisassemblyOfControlCabinets}. Similarly, Zamora-Hernández et al. combine YOLO-based object detection with action recognition to infer executed assembly steps from image sequences and compare them against a predefined process model represented as a state machine \cite{DeepLearningBasedVisualControlAssistent}. Basamakis et al. extend this paradigm by integrating object detection and quality inspection with a hybrid pose estimation pipeline that combines convolutional neural networks with classical computer vision techniques. To facilitate training, their approach exploits both real and synthetic data generated from CAD models \cite{DeepLearningAdvancedPoseEstimation}.

Beyond single-image analysis, several researchers have incorporated temporal information from video recordings. Chen et al. employ convolutional neural networks to classify assembly actions from video sequences while simultaneously performing semantic segmentation of relevant components. Both tasks are trained in a fully supervised manner using manually annotated datasets \cite{MonitoringAssemblyProcessUsingDeepLearning}.

To overcome the limitations associated with predefined component classes, representation learning approaches have recently gained attention. Schoonbeek et al. propose a framework conceptually related to SimCLR, in which images depicting similar assembly states are encouraged to produce similar embeddings, while embeddings of different states are pushed apart. This enables the model to focus on assembly-relevant visual characteristics rather than memorizing specific component identities, thereby improving generalization across assembly scenarios \cite{TowerdsGeneralizableAssemblyStateRecognition}.

In contrast to these methods, JEPA-based approaches learn representations in a self-supervised manner and therefore offer the potential to reduce the dependence on labeled training data while generalizing beyond fixed sets of component classes. Unlike most computer-vision applications of JEPAs, which retain only the encoder, JEPA-based world models also use the predictor to model state transitions in latent space. For assembly verification, this allows the expected post-action state to be predicted and compared with the observed one.

To the best of our knowledge, the application of action-conditioned JEPA world models to fine-grained assembly-step verification has not yet been investigated. Given their ability to learn action-dependent state transitions in a self-supervised manner, they represent a promising approach for industrial assembly environments characterized by frequent product changes, substantial visual variability, and limited availability of annotated data.

\section{Background}
This section introduces the theoretical foundations of JEPA-based world models relevant to our work. V-JEPA, introduced by Bardes et al., demonstrated that semantically rich visual representations can be learned from video without relying on annotations or labels \cite{v-jepa}. The model employs a masking strategy conceptually related to those used in Masked Autoencoders \cite{MaskedAutoencoders} and I-JEPA \cite{IJepa}. During training, spatial and temporal regions of the input video are masked and provided to a context encoder, while a target encoder processes the corresponding unmasked video. Based on the representations produced by the context encoder, the predictor learns to predict the target-encoder representations corresponding to the masked regions.
To prevent a trivial solution in which both encoders collapse to a constant representation, the weights of the target encoder are updated using an exponential moving average (EMA) of the context encoder's weights.

Building on this work, Garrido et al. showed that an understanding of intuitive physics can emerge through self-supervised pretraining on natural videos \cite{garrido2025intuitivephysicsunderstandingemerges}. They evaluated V-JEPA on both physically plausible and implausible videos by measuring the discrepancy between predicted representations and those of the observed future frames. When physically implausible events occur, such as objects unexpectedly disappearing, the prediction error increases, whereas it remains comparatively low for physically consistent events.

This property is also central to our work, as it provides a framework for determining whether an observed state transition deviates from the state transitions learned during training. We aim to exploit this principle to learn a world model of assembly processes and use the resulting prediction error for assembly step verification. 

This requires a world model that predicts state transitions conditioned on the action being performed. LeWorldModel \cite{maes2026leworldmodelstableendtoendjointembedding} extends the JEPA-based world model paradigm in this direction and serves as the architectural foundation for the world model developed in this work. 

A key difference from earlier JEPA-based approaches is that representation collapse is prevented without relying on training mechanisms such as stop-gradient operations or exponential moving average (EMA) updates of the target encoder. Instead, LeWorldModel introduces an additional regularization term, called Sketched Isotropic Gaussian Regularization (SIGReg), into the training objective. This regularization encourages the encoder representations to follow an isotropic Gaussian distribution in the latent space, thereby preventing the encoder from collapsing to a constant representation.

LeWorldModel learns the dynamics of the environment directly from sequences of image observations and actions. Given the representation of the current observation and the corresponding action, the predictor learns to predict the latent representation of the subsequent state. The overall architecture and the interaction between the encoder, predictor, and SIGReg regularization are illustrated in Fig.~\ref{fig:leworldmodel}. In this work, we build on this action-conditioned prediction principle and use the discrepancy between the predicted and observed subsequent state as the basis for assembly step verification.

\begin{figure*}[t] \centering \begin{tabular}{c >{\columncolor{gray!15}}c >{\columncolor{gray!15}}c >{\columncolor{gray!15}}c >{\columncolor{blue!6}}c} & \multicolumn{3}{>{\columncolor{gray!15}}c}{\textbf{Control Cabinets Dataset}} & \cellcolor{blue!6}\textbf{Modified CLEVR-Change} \\ & \textbf{Slotted Screw} & \textbf{Washer} & \textbf{Nut} & \textbf{Material Change} \\ \raisebox{5.5\height}{\textbf{Before}} & \includegraphics[width=0.2\textwidth]{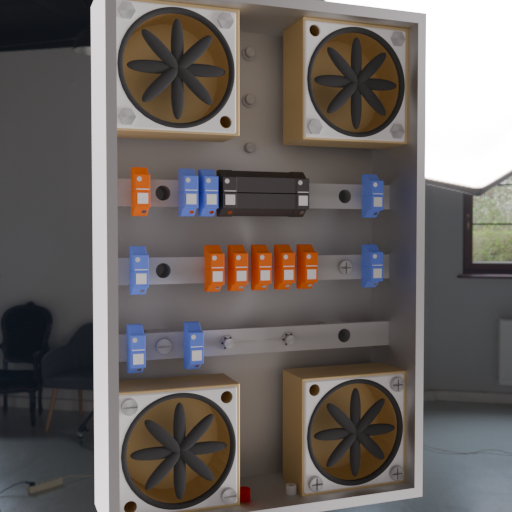} & \includegraphics[width=0.2\textwidth]{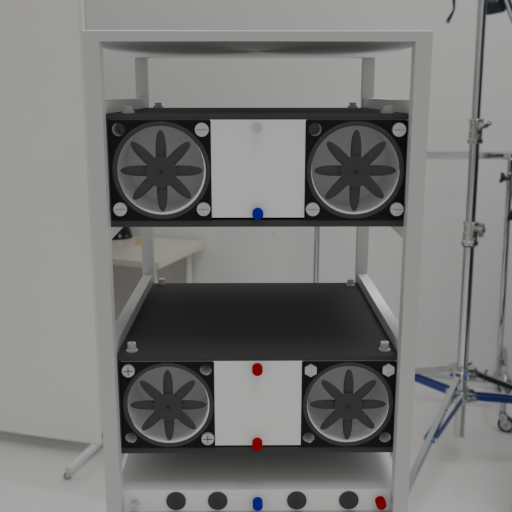} & \includegraphics[width=0.2\textwidth]{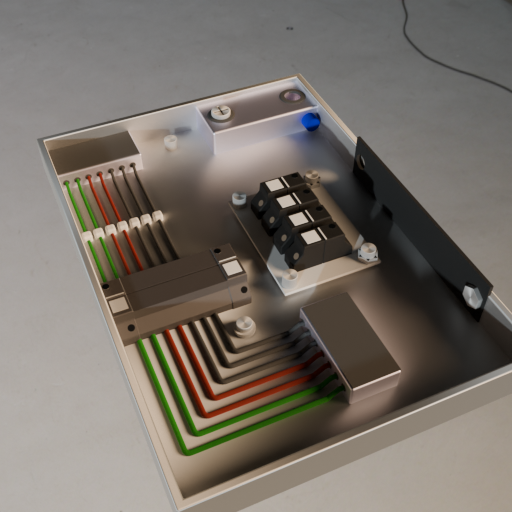} & \includegraphics[width=0.2\textwidth]{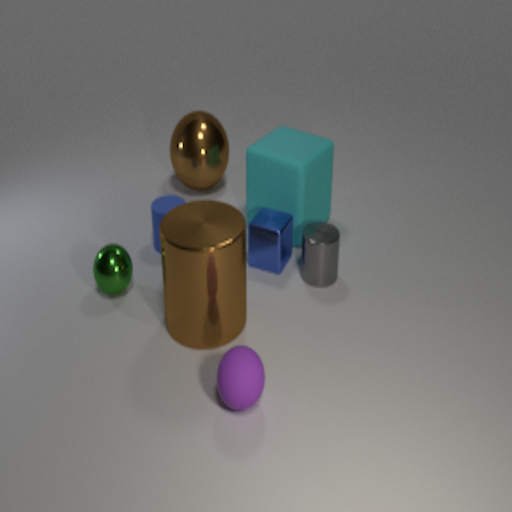} \\[4pt] \raisebox{5.5\height}{\textbf{After}} & \includegraphics[width=0.2\textwidth]{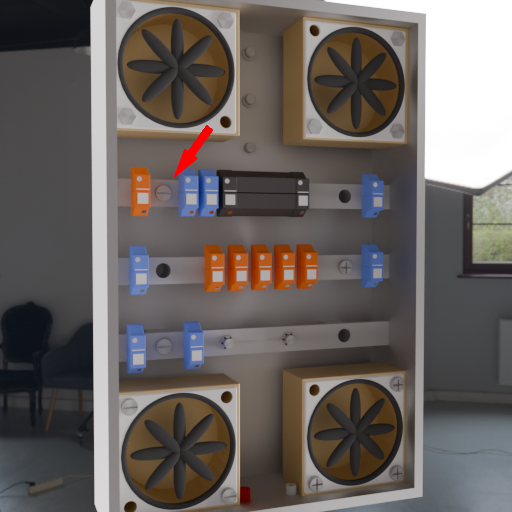} & \includegraphics[width=0.2\textwidth]{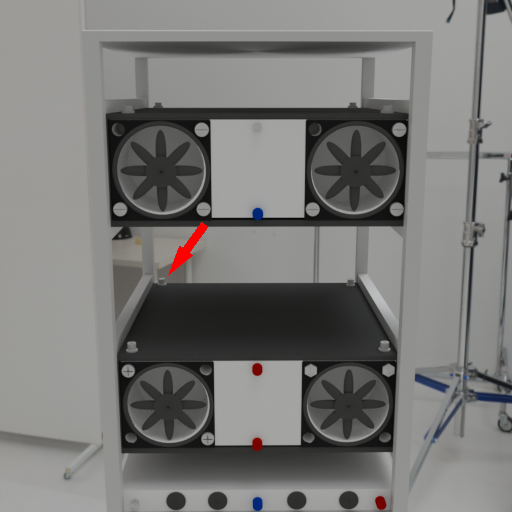} & \includegraphics[width=0.2\textwidth]{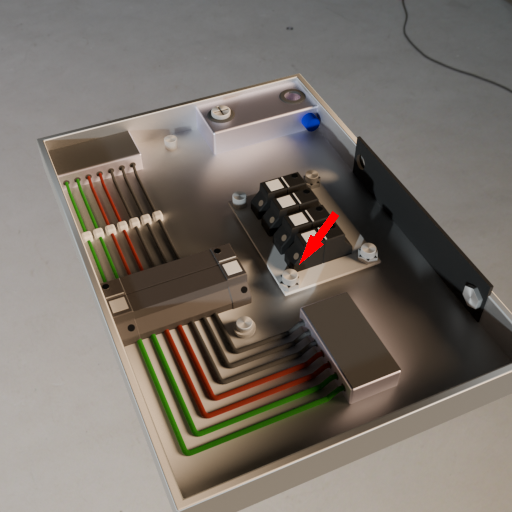} & \includegraphics[width=0.2\textwidth]{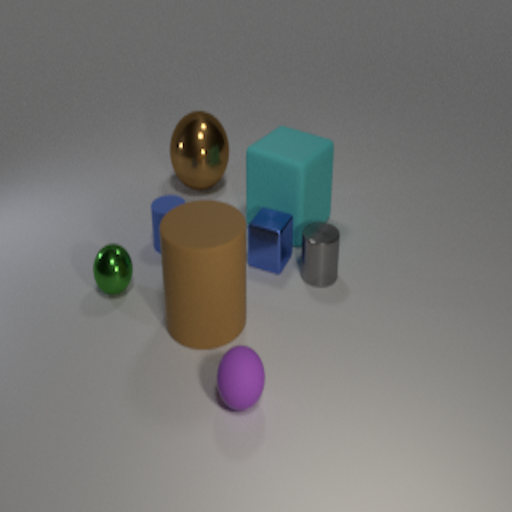} \\ \end{tabular} \caption{Example image pairs from the datasets. The first three columns show examples from the Control Cabinets Dataset, depicting distinct assembly actions together with the corresponding assembly states before and after execution. The red arrow indicates the location where the assembly action was performed. The last column shows an example from the modified CLEVR-Change dataset, illustrating a material change of the central cylinder between the two observations.} \label{fig:dataset_examples} \end{figure*}

\section{Problem Definition}
To verify the correctness of assembly steps, we formulate the verification problem as 
\begin{equation} 
    d = 1 - \cos\!\left( p\!\left(e(\mathbf{x}_t), \mathbf{a}_t\right), e(\mathbf{x}_{t+1}) \right), 
    \label{eq:verification_distance} 
\end{equation} where $\mathbf{x}_t$ represents the image before and $\mathbf{x}_{t+1}$ the image after the assembly step. Both images are encoded by the encoder $e(\cdot)$. The assembly operation is represented by the discrete action $\mathbf{a}_t$ and, together with the embedding of $\mathbf{x}_t$, serves as input to the predictor $p(\cdot)$. A correctly executed assembly step should result in a target embedding $e(\mathbf{x}_{t+1})$ that closely matches the predictor output $p(e(\mathbf{x}_t), \mathbf{a}_t)$, yielding a cosine distance close to zero, \begin{equation} d \rightarrow 0 \quad \text{for valid assembly transitions.} \end{equation} In contrast, incorrect assembly states or mismatched actions are expected to produce substantially larger distances, 

\begin{equation} 
    d \rightarrow 2 \quad \text{for invalid assembly transitions.} 
\end{equation}

\section{Dataset}
\label{sec:Dataset}
To match our problem definition, the dataset must consist of image pairs depicting the same scene before and after an assembly step, together with a discrete action label. While not a strict requirement of the underlying problem, our approach assumes a largely consistent camera viewpoint across observations, a static object that remains stationary during the assembly process, and a background free from unrelated changes. These assumptions enable the model to focus on assembly-induced changes rather than compensating for irrelevant visual variations caused by camera motion, viewpoint shifts, background changes, or movement of the assembly object itself.

Since no publicly available data satisfied these requirements, we created two synthetic datasets:

\paragraph{Control Cabinets} Our first custom dataset comprises 10 control cabinets, 7 distinct assembly actions, and 162 assembly positions. For each assembly position, $512 \times 512$ pixel images were rendered from 20 viewpoints under 5 different lighting conditions, resulting in a total of 16,200 image pairs.

We split the dataset at the control cabinet level to evaluate generalization to previously unseen assemblies. The first eight control cabinets are used for training, the ninth for validation, and the tenth for testing. 

To evaluate whether the model can detect assembly errors, we generate negative samples corresponding to two distinct failure modes. In the first scenario, referred to as \textit{wrong action}, the post-assembly image remains unchanged while the action label is replaced by a randomly selected different action, simulating the execution of an incorrect assembly step. In the second scenario, referred to as \textit{wrong state}, the action label remains unchanged, but the post-assembly image is replaced with the corresponding pre-assembly image, simulating a missing assembly step. In both cases, the action-conditioned prediction is expected to be inconsistent with the provided post-assembly image, resulting in a large distance between the predicted representation and the representation of the post-assembly image. Example images from the dataset are shown in Fig.~\ref{fig:dataset_examples}.

\paragraph{Modified CLEVR-Change} 
In addition, we created a custom version of the publicly available CLEVR-Change dataset \cite{ClevrChange}. The dataset consists of before-and-after image pairs depicting the same scene, where a single object attribute changes between both observations. Unlike the original dataset, all scenes were re-rendered with a fixed camera viewpoint to eliminate appearance changes caused by camera motion. The considered modifications include changes in object color, changes in material, as well as object insertions and removals. Overall, the dataset contains 20,000 image pairs and 12 distinct actions describing the transformation from the before image to the after image. An example image pair from the dataset is shown in Fig.~\ref{fig:dataset_examples}.

\begin{figure*}[t] 
\centering 
\includegraphics[width = 420pt]{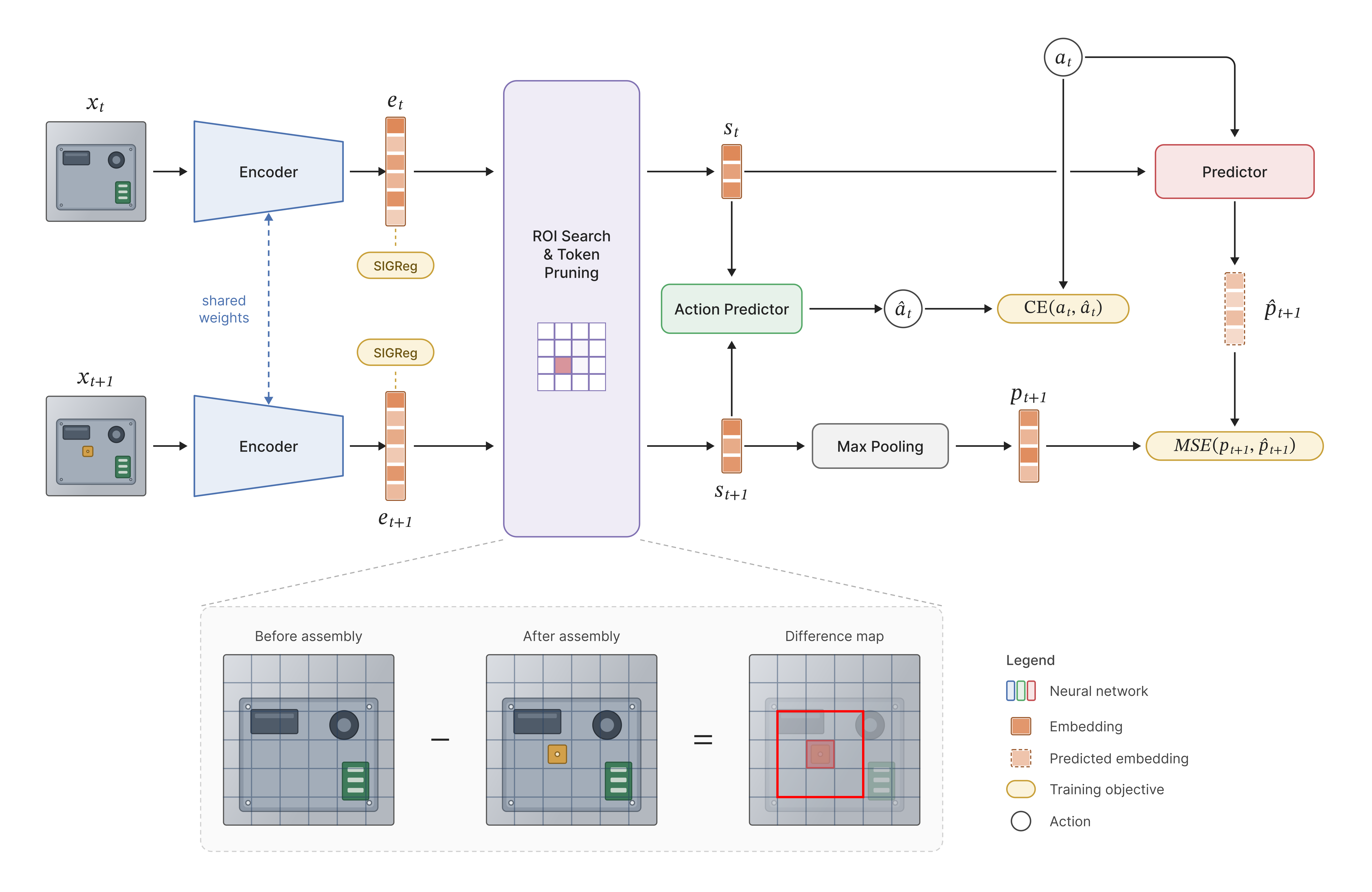} 
\caption{Overview of the proposed Assembly-JEPA architecture. Images captured before and after an assembly step are encoded into patch-level representations using a shared encoder. A difference map is computed from the pre- and post-action patch embeddings to localize the assembly-relevant region, from which a subset of embeddings is extracted and processed by an action-conditioned predictor. To encourage action-aware representations, an auxiliary inverse action prediction objective is applied to the latent differences between pre- and post-action embeddings. Verification is performed by comparing the predicted post-action representation with the observed representation of the post-action image.} 
\label{fig:AssemblyJepaArchitecture}
\end{figure*}

\section{Method: Assembly-JEPA for Image Verification} 
Modern world models such as V-JEPA \cite{v-jepa} have demonstrated not only the ability to learn highly informative feature representations from video data, but also their capability to detect implausible or physically impossible events \cite{garrido2025intuitivephysicsunderstandingemerges}. By predicting the representation of future frames from past observations and measuring the prediction error between the predicted representation and the actual representation of the future frame, it becomes possible to identify implausible behavior. Plausible and physically consistent sequences generally result in a low prediction error, whereas implausible sequences lead to a significantly higher prediction error. We aim to exploit this property to address a fine-grained industrial image verification problem. 

Given an image captured before an assembly step, our goal is to use a JEPA model conditioned on the corresponding assembly action to predict the representation of the image after the assembly step. By measuring the prediction error between the predicted representation and the actual representation of the post-assembly image, we can determine whether the assembly step was executed correctly. 

As the foundation of our approach, we adopt LeWorldModel and adapt it to our specific task \cite{maes2026leworldmodelstableendtoendjointembedding}. A key difference compared to the original application domain of LeWorldModel is that our approach operates on before-and-after assembly images rather than videos and must be discriminative with respect to fine-grained visual changes between these images. 

We evaluate the verification performance using the area under the receiver operating characteristic curve (AUROC). Since the proposed method assigns a continuous verification score to each before-and-after image pair, AUROC allows us to assess how well the model discriminates between correctly executed assembly operations (positive samples) and incorrectly or non-executed operations (negative samples) across all possible decision thresholds. This avoids selecting a task-specific decision threshold solely for evaluation.

When applied to our assembly verification task without any task-specific modifications, the original LeWorldModel architecture achieves a validation AUROC of only 50.02\%, corresponding to near-random performance. This result indicates that state-of-the-art JEPA-based world models are not directly applicable to fine-grained assembly verification and motivates the design choices proposed in the following sections.

\subsection{Architecture} 
\label{sec:Architecture}

Fine-grained assembly verification poses two main challenges. First, individual assembly actions may affect only small and visually subtle image regions. Second, the resulting state transition is spatially non-deterministic, as the same assembly action may be performed at different locations. To address these characteristics, Assembly-JEPA operates on patch-level embeddings and localizes the region most relevant to the observed state transition. In addition, an auxiliary inverse action prediction objective encourages the encoder to preserve action-specific information, while max pooling reduces the influence of unchanged patch representations on the prediction objective. The complete architecture is illustrated in Fig.~\ref{fig:AssemblyJepaArchitecture}.

\subsubsection{Patch-based Encoder} To better capture fine-grained changes between the pre- and post-assembly images, the single CLS token representation is replaced with patch embeddings:

\begin{equation} 
\mathbf{e}_{t}, \mathbf{e}_{t+1} \in \mathbb{R}^{H \times W \times D}
\end{equation} 

where $\mathbf{e}_{t}$ and $\mathbf{e}_{t+1}$ denote the patch embeddings extracted from the images before and after the assembly step, respectively. The embedding dimension is set to $D = 64$ in all experiments.

Furthermore, we replace the ViT-Tiny encoder used in LeWorldModel with a lightweight ResNet18 and interpret the encoder's final feature map as a grid of patch embeddings, where each spatial location corresponds to an image region represented by a $D$-dimensional feature vector. 

\subsubsection{Action-conditioned Predictor} Following LeWorldModel, we use a Transformer-based predictor conditioned on assembly actions via Adaptive Layer Normalization (AdaLN). The predictor comprises 8 Transformer blocks and 16 attention heads.

\subsubsection{ROI Localization} To address the non-deterministic nature of the task, we exploit a key difference between our setting and the classical world-model setting, namely that verification is performed after the assembly action has already been executed. Consequently, both the before and after image are available during inference, allowing us to search for the location at which the assembly action was performed. 

\begin{equation} 
\label{eq:crop} 
\begin{gathered} 
\Delta_{i,j} = \left\| \mathbf{e}_{t+1,i,j} - \mathbf{e}_{t,i,j} \right\|_2, \\ 
(i^{*},j^{*}) = \arg\max_{(i,j)} \Delta_{i,j}, \\ 
\mathbf{s}_{t+1} = \operatorname{Crop}\!\left(\mathbf{e}_{t+1}, i^*, j^*, n, m\right), \\ 
\mathbf{s}_{t} = \operatorname{Crop}\!\left(\mathbf{e}_{t}, i^*, j^*, n, m\right), \\ 
\mathbf{s}_{t+1}, \mathbf{s}_{t} \in \mathbb{R}^{n \times m \times D} 
\end{gathered} 
\end{equation}

The first step computes a difference map from the patch embeddings of the before and after images and identifies the location with the highest difference value. Assuming that the patch embeddings capture the semantic content of the image, the location with the highest difference corresponds to the image region that changed most significantly and therefore likely indicates where the assembly action was performed. This localization assumes that the change is concentrated within a single spatial region.

We then extract an $n \times m$ token window centered around the location with the highest difference value and use the resulting tokens for subsequent processing, yielding the final patch embeddings defined in \eqref{eq:crop}. This not only resolves the non-deterministic nature of the task but also further supports fine-grained discrimination by allowing the predictor to focus exclusively on the region relevant to the assembly action. In all experiments, we use a fixed window size of $n=m=5$.

\subsubsection{Inverse Action Prediction}

Although ROI localization restricts the prediction task to the action-relevant image region, spatial localization alone does not ensure that the encoder preserves information about the performed assembly action. This is particularly important for visually subtle state transitions, where the before and after representations may differ only marginally. To explicitly encourage action-sensitive latent representations, we adopt an idea proposed by Zhang et al., who predict actions from differences between encoder representations \cite{zhang2026deltajepalearningactionsensitiveworld}. Specifically, we apply:

\begin{equation} 
\begin{aligned} 
\mathbf{p}_{t+1} &= \operatorname{MaxPool}(\mathbf{s}_{t+1}), \\ 
\mathbf{p}_t &= \operatorname{MaxPool}(\mathbf{s}_t), \\ 
\mathbf{\Delta s} &= \mathbf{p}_{t+1} - \mathbf{p}_t, \\ 
\hat{\mathbf{a}}_t &= W\mathbf{\Delta s} + b, \\ 
\mathcal{L}_{\mathrm{iap}} &= \operatorname{CE}\!\left(\hat{\mathbf{a}}_t, \mathbf{a}_t\right). 
\end{aligned} 
\end{equation}

where $\mathbf{s}_{t}$ and $\mathbf{s}_{t+1}$ denote the cropped patch embeddings, $W$ is a trainable weight matrix, $b$ is a trainable bias term, $\mathbf{a}_t$ is the one-hot encoded action vector describing the assembly action performed between the before and after image, and $\operatorname{CE}$ denotes the cross-entropy loss.

We deliberately use only a single linear layer for action prediction. This design encourages the encoder to learn a latent space in which representations associated with different actions, as well as representations from samples without a visible assembly action, are linearly separable.

The final training objective combines the original LeWorldModel prediction and regularization objectives with the proposed inverse action prediction loss:
\begin{equation} \mathcal{L}_{\mathrm{total}} = (1-\lambda)\,\mathcal{L}_{\mathrm{pred}} + \lambda\,\mathcal{L}_{\mathrm{sigreg}} + \mathcal{L}_{\mathrm{iap}},
\end{equation}
where $\mathcal{L}_{\mathrm{pred}}$ denotes the prediction loss, $\mathcal{L}_{\mathrm{sigreg}}$ the Sketched Isotropic Gaussian regularization loss \cite{balestriero2025lejepaprovablescalableselfsupervised}, and $\mathcal{L}_{\mathrm{iap}}$ the cross-entropy loss used for the inverse action prediction.

\subsubsection{Max-Pooling of Patch Embeddings}
Even after restricting the embeddings to the relevant image region, many assembly actions affect only a very small subset of patch embeddings, often a single patch. Consequently, most target embeddings remain nearly unchanged between the before and after image. This allows the predictor to solve large parts of the task through trivial identity mappings instead of focusing on the assembly-induced changes. 

To mitigate this issue, we modify the prediction target by aggregating the target patch embeddings using a max-pooling operation across the sequence dimension: 

\begin{equation} \mathbf{p}_{t+1} = \operatorname{MaxPool}(\mathbf{s}_{t+1}), \qquad \mathbf{p}_{t+1} \in \mathbb{R}^{D}. \end{equation}

On the predictor side, a single query token is appended to the input representation, and the corresponding output token is trained to predict the pooled target vector $\mathbf{p}_{t+1}$. By reducing the target representation to a single embedding, the model is encouraged to focus on the most salient assembly-induced changes rather than reconstructing a large number of largely unchanged patch embeddings. 

The resulting Assembly-JEPA architecture achieves a validation AUROC of 88.5\% on the control cabinets dataset. The contribution of the individual components is investigated through the ablation study presented in the following section.

\begin{figure}[b] \centering \includegraphics[width=200pt]{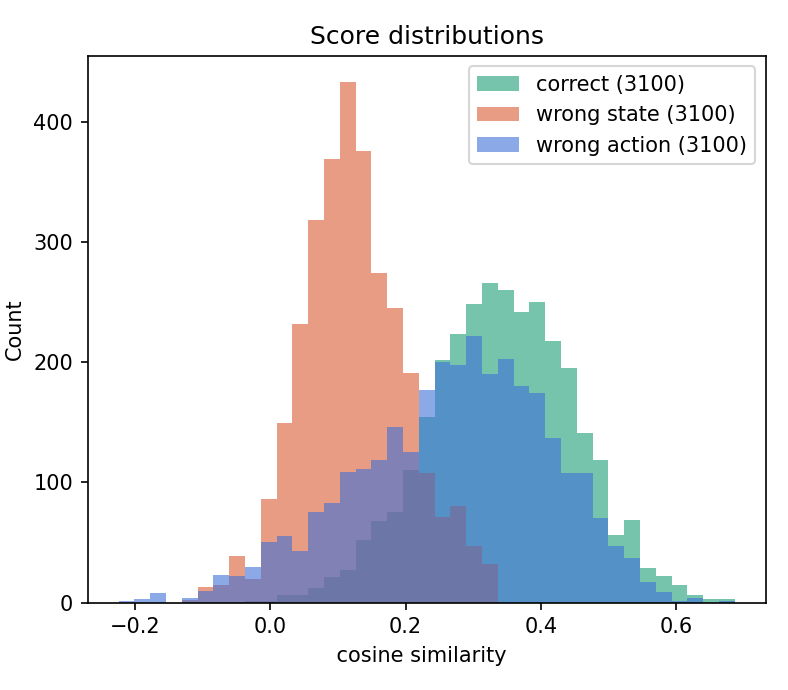} \caption{Distribution of verification scores for positive, wrong state, and wrong action samples using the ROI-localization-only configuration.} \label{fig:score_histogram} \end{figure}

\subsection{Ablation Study}

\begin{table*}[t] \centering \caption{Ablation study of the proposed Assembly-JEPA components. The hard wrong-state setting retains the correct assembly region when constructing negative samples, resulting in a more challenging evaluation of fine-grained state transitions.} \label{tab:ablation_study} \scriptsize \setlength{\tabcolsep}{4pt} \renewcommand{\arraystretch}{1.2} \begin{tabularx}{\textwidth}{ >{\centering\arraybackslash}p{0.11\textwidth} >{\centering\arraybackslash}p{0.13\textwidth} >{\centering\arraybackslash}p{0.10\textwidth} >{\centering\arraybackslash}p{0.14\textwidth} >{\centering\arraybackslash}p{0.07\textwidth} >{\raggedright\arraybackslash}X } \toprule \textbf{ROI localization} & \textbf{Inverse action prediction} & \textbf{Max pooling} & \textbf{Evaluation setting} & \textbf{Val. AUROC} & \textbf{Purpose} \\ \midrule \cmark & \xmark & \xmark & Standard & 79.5\% & Initial configuration demonstrating the effect of localizing the action-relevant image region. \\ \cmark & \xmark & \xmark & Hard wrong state & 59.9\% & Evaluation on hard wrong-state samples, revealing that ROI localization alone does not reliably capture fine-grained state transitions. \\ \cmark & \cmark & \xmark & Hard wrong state & 76.1\% & Addition of the inverse action prediction loss to improve the representation of fine-grained assembly state transitions. \\ \cmark & \cmark & \cmark & Hard wrong state & 81.3\% & Addition of max pooling to reduce the influence of unchanged patch embeddings on the prediction objective. \\ \cmark & \cmark & \cmark & Standard & 88.5\% & Final Assembly-JEPA architecture evaluated on the standard verification benchmark. \\ \bottomrule \end{tabularx} \end{table*}

To assess the contribution of the individual components of Assembly-JEPA, we conduct a series of ablation experiments summarized in Table~\ref{tab:ablation_study}. In addition to the standard verification benchmark, we introduce a more challenging wrong-state setting to investigate whether the model captures fine-grained state transitions or exploits differences introduced by the ROI localization procedure. The ablation therefore serves two purposes: first, to analyze the limitations of ROI localization in isolation, and second, to quantify the contributions of inverse action prediction and max pooling to the final architecture.

\subsubsection{Analysis of ROI Localization}
The ROI-localization-only configuration achieves an AUROC of 79.5\% on the standard validation benchmark. However, the verification score distributions in Fig.~\ref{fig:score_histogram} reveal a substantial difference between the two types of negative samples. While wrong action samples strongly overlap with positive samples, wrong state samples exhibit a more distinct score distribution. This behavior can partially be attributed to the construction of wrong state samples. If no change is detected between the images used for ROI localization, the localization procedure falls back to a predefined reference region. 
Consequently, the resulting embeddings may originate from an image region unrelated to the actual assembly operation.

\begin{figure*}[t] 
\centering 
\includegraphics[width=1.0\textwidth]{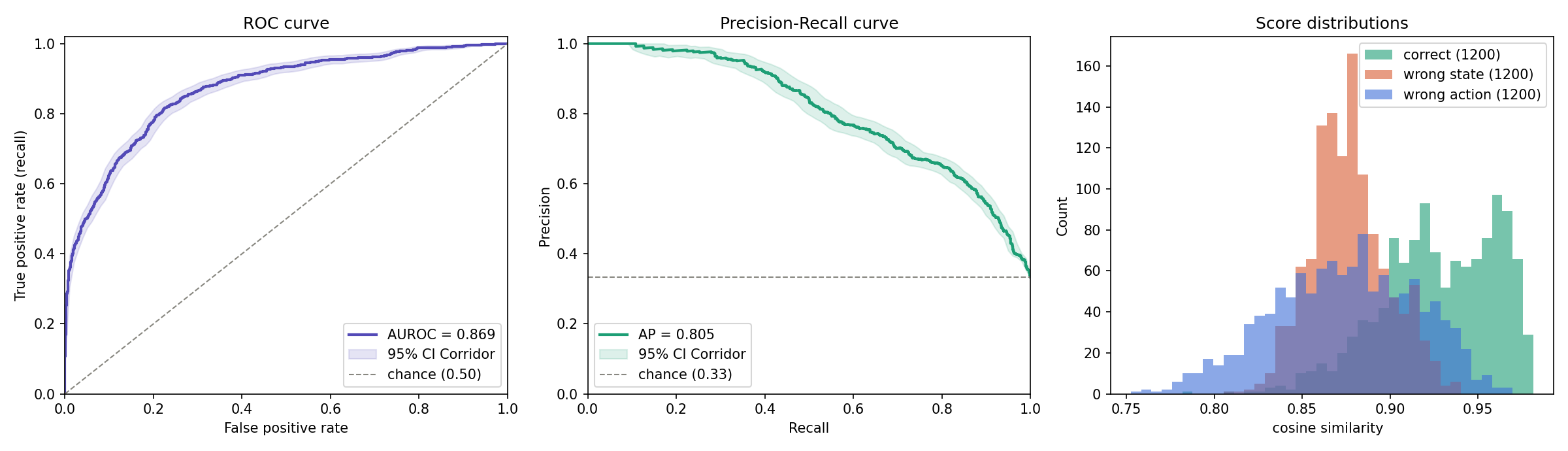} 
\vspace{0.5em} 
\includegraphics[width=1.0\textwidth]{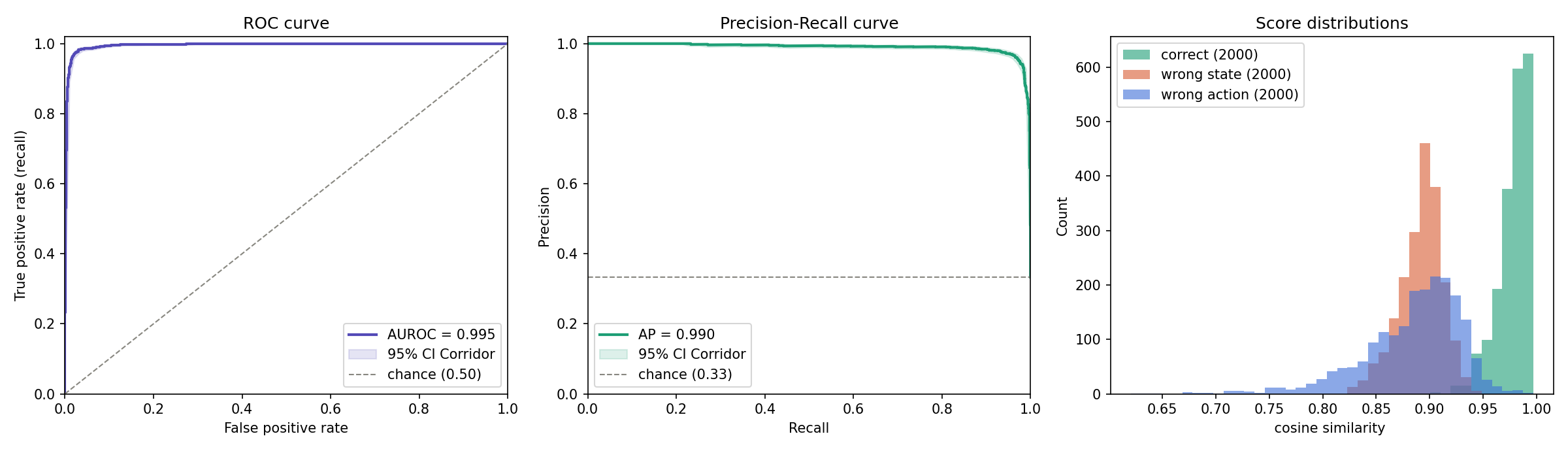} 
\caption{ Comparison of Assembly-JEPA on the assembly dataset (top) and the modified CLEVR-Change dataset (bottom), demonstrating the model's ability to distinguish between valid and invalid state transitions. On the modified CLEVR-Change dataset, the model achieves a near-perfect AUROC of 99.5\%, with almost complete separation between positive and negative samples. The substantially higher performance on CLEVR-Change can be attributed to the more pronounced visual state changes present in the dataset, whereas the proposed assembly benchmark requires distinguishing highly localized and visually subtle assembly transitions. Shaded regions denote 95\% bootstrap confidence intervals. } 
\label{fig:EvaluationComparison} 
\end{figure*}

To isolate this effect, we construct a more challenging wrong state benchmark in which ROI localization is first performed using the original image pair. Only after the assembly-relevant region has been localized are the post-action embeddings replaced by the corresponding pre-action embeddings. This preserves the actual assembly location while representing a non-executed action.

When evaluated under this more challenging setting, the ROI-localization-only configuration achieves an AUROC of 59.9\%, compared with 79.5\% on the standard benchmark, indicating that ROI localization alone is insufficient for distinguishing subtle executed and non-executed state transitions.

\subsubsection{Contribution of Inverse Action Prediction}
Adding inverse action prediction substantially improves performance on the challenging wrong state benchmark, increasing AUROC from 59.9\% to 76.1\%. This corresponds to an absolute improvement of 16.2 percentage points over ROI localization alone under the same evaluation setting. This improvement supports the intended effect of the inverse action prediction objective, encouraging latent representations that are more sensitive to fine-grained assembly state transitions.

\subsubsection{Contribution of Max Pooling}
Combining inverse action prediction with the max-pooled prediction target further increases AUROC to 81.3\% on the challenging benchmark. This represents a further improvement of 5.2 percentage points over inverse action prediction without max pooling, supporting the motivation for reducing the influence of unchanged patch embeddings on the prediction objective.

On the standard benchmark, the complete Assembly-JEPA architecture achieves an AUROC of 88.5\%, compared with 79.5\% for ROI localization alone. Overall, the ablation results show that ROI localization provides a strong initial improvement, while inverse action prediction and max pooling address its limitations in representing subtle assembly state transitions and contribute to the performance of the final architecture.

\section{Evaluation}

\subsection{Evaluation on Custom Assembly Dataset} 
We first evaluate the proposed approach by training on the first eight control cabinets and testing on the held-out tenth control cabinet. Unless stated otherwise, all models are trained for 50 epochs using the AdamW optimizer with a learning rate of $1 \times 10^{-4}$ and a batch size of 64. For each test sample, the cosine similarity between the predicted and target embedding is computed.

Performance is quantified using ROC and Precision-Recall curves, and confidence intervals are estimated using bootstrap resampling. Valid assembly transitions are treated as positive samples, while the wrong action and wrong state samples introduced in Section \ref{sec:Dataset} serve as negative samples. 

As shown in Fig.~\ref{fig:EvaluationComparison}, the proposed approach generalizes to previously unseen control cabinets under a cabinet-level train-test split. While some overlap between valid and invalid transitions remains visible in the score distributions, correct assembly transitions are generally associated with higher similarity values than wrong-state and wrong-action samples. This indicates that the learned latent representations capture relevant information about assembly state changes and can be used to distinguish valid from invalid assembly outcomes without requiring explicit defect examples during training.

\subsection{Evaluation on Modified CLEVR-Change}

In addition to our assembly dataset, we evaluated the proposed method on the modified CLEVR-Change dataset using the same training procedure as in the assembly experiments. On the CLEVR-Change test set, the model reached a near-perfect AUROC of 99.5\%, as illustrated in Fig.~\ref{fig:EvaluationComparison}. This result demonstrates that the proposed approach generalizes beyond our custom assembly dataset. The substantially higher performance compared to the assembly dataset can be attributed to the more pronounced state changes present in CLEVR-Change, whereas industrial assembly verification requires distinguishing considerably more subtle visual differences. Interestingly, this large performance gap suggests that benchmark difficulty is strongly influenced by the granularity of the underlying state changes. While CLEVR-Change primarily contains clearly visible object-level modifications, the assembly dataset requires distinguishing highly localized and visually subtle state transitions, making it a substantially more challenging evaluation scenario for action-conditioned state verification.

\section{Conclusion} 
In this paper, we introduced Assembly-JEPA, a novel JEPA-based model for fine-grained industrial image verification. Unlike previous JEPA applications in computer vision, our approach retains the complete architecture, including the predictor, during inference and leverages action-conditioned latent-space prediction for assembly-state verification. Through a series of targeted architectural modifications, we demonstrated that the model can learn fine-grained state transitions and detect subtle assembly deviations in a self-supervised manner without requiring explicit negative samples. 

Our experiments further highlight that generic predictive objectives alone may be insufficient for fine-grained industrial verification tasks. In particular, the results show that explicitly encouraging the encoder to capture action-specific information substantially improves the model's ability to distinguish visually similar assembly states. 

While the proposed approach achieves promising results on both our custom assembly dataset and the modified CLEVR-Change benchmark, further evaluation on real-world industrial data is required to assess its practical applicability and robustness. In addition, the current ROI localization strategy assumes that the most significant visual change corresponds to the executed assembly action. Future work could therefore focus on more robust localization mechanisms that remain reliable under viewpoint changes, illumination variations, and other irrelevant scene modifications.

\clearpage
\bibliographystyle{IEEEtran}
\bibliography{bibliography}

\begin{thebibliography}{10}
\providecommand{\url}[1]{#1}
\csname url@samestyle\endcsname
\providecommand{\newblock}{\relax}
\providecommand{\bibinfo}[2]{#2}
\providecommand{\BIBentrySTDinterwordspacing}{\spaceskip=0pt\relax}
\providecommand{\BIBentryALTinterwordstretchfactor}{4}
\providecommand{\BIBentryALTinterwordspacing}{\spaceskip=\fontdimen2\font plus
\BIBentryALTinterwordstretchfactor\fontdimen3\font minus \fontdimen4\font\relax}
\providecommand{\BIBforeignlanguage}[2]{{%
\expandafter\ifx\csname l@#1\endcsname\relax
\typeout{** WARNING: IEEEtran.bst: No hyphenation pattern has been}%
\typeout{** loaded for the language `#1'. Using the pattern for}%
\typeout{** the default language instead.}%
\else
\language=\csname l@#1\endcsname
\fi
#2}}
\providecommand{\BIBdecl}{\relax}
\BIBdecl

\bibitem{asi7010011}
\BIBentryALTinterwordspacing
N.~Hütten, M.~Alves~Gomes, F.~Hölken, K.~Andricevic, R.~Meyes, and T.~Meisen, ``Deep learning for automated visual inspection in manufacturing and maintenance: A survey of open-access papers,'' \emph{Applied System Innovation}, vol.~7, no.~1, 2024, 10.3390/asi7010011. [Online]. Available: \url{https://www.mdpi.com/2571-5577/7/1/11}
\BIBentrySTDinterwordspacing

\bibitem{ResNet}
K.~He, X.~Zhang, S.~Ren, and J.~Sun, ``Deep residual learning for image recognition,'' in \emph{2016 IEEE Conference on Computer Vision and Pattern Recognition (CVPR)}, 2016, pp. 770--778, 10.1109/CVPR.2016.90.

\bibitem{Yolo}
\BIBentryALTinterwordspacing
J.~Redmon, S.~Divvala, R.~Girshick, and A.~Farhadi, ``{You Only Look Once: Unified, Real-Time Object Detection},'' in \emph{2016 IEEE Conference on Computer Vision and Pattern Recognition (CVPR)}.\hskip 1em plus 0.5em minus 0.4em\relax Los Alamitos, CA, USA: IEEE Computer Society, Jun. 2016, pp. 779--788, 10.1109/CVPR.2016.91. [Online]. Available: \url{https://doi.ieeecomputersociety.org/10.1109/CVPR.2016.91}
\BIBentrySTDinterwordspacing

\bibitem{Padim}
T.~Defard, A.~Setkov, A.~Loesch, and R.~Audigier, ``{PaDiM}: A patch distribution modeling framework for anomaly detection and localization,'' in \emph{Pattern Recognition. ICPR International Workshops and Challenges}, A.~Del~Bimbo, R.~Cucchiara, S.~Sclaroff, G.~M. Farinella, T.~Mei, M.~Bertini, H.~J. Escalante, and R.~Vezzani, Eds.\hskip 1em plus 0.5em minus 0.4em\relax {Cham, Switzerland}: Springer International Publishing, 2021, pp. 475--489, 10.1007/978-3-030-68799-1\_35.

\bibitem{PatchCore}
K.~Roth, L.~Pemula, J.~Zepeda, B.~Schölkopf, T.~Brox, and P.~Gehler, ``Towards total recall in industrial anomaly detection,'' in \emph{2022 IEEE/CVF Conference on Computer Vision and Pattern Recognition (CVPR)}, 2022, pp. 14\,298--14\,308, 10.1109/CVPR52688.2022.01392.

\bibitem{MVTec}
P.~Bergmann, M.~Fauser, D.~Sattlegger, and C.~Steger, ``{MVTec AD} — a comprehensive real-world dataset for unsupervised anomaly detection,'' in \emph{2019 IEEE/CVF Conference on Computer Vision and Pattern Recognition (CVPR)}, 2019, pp. 9584--9592, 10.1109/CVPR.2019.00982.

\bibitem{BeyondAcademicBenchmarks}
\BIBentryALTinterwordspacing
A.~Baitieva, Y.~Bouaouni, A.~Briot, D.~Ameln, S.~Khalfaoui, and S.~Akcay, ``{Beyond Academic Benchmarks: Critical Analysis and Best Practices for Visual Industrial Anomaly Detection},'' in \emph{2025 IEEE/CVF Conference on Computer Vision and Pattern Recognition Workshops (CVPRW)}.\hskip 1em plus 0.5em minus 0.4em\relax Los Alamitos, CA, USA: IEEE Computer Society, Jun. 2025, pp. 4015--4025, 10.1109/CVPRW67362.2025.00386. [Online]. Available: \url{https://doi.ieeecomputersociety.org/10.1109/CVPRW67362.2025.00386}
\BIBentrySTDinterwordspacing

\bibitem{IJepa}
M.~Assran, Q.~Duval, I.~Misra, P.~Bojanowski, P.~Vincent, M.~Rabbat, Y.~LeCun, and N.~Ballas, ``Self-supervised learning from images with a joint-embedding predictive architecture,'' in \emph{2023 IEEE/CVF Conference on Computer Vision and Pattern Recognition (CVPR)}, 2023, pp. 15\,619--15\,629, 10.1109/CVPR52729.2023.01499.

\bibitem{balestriero2025lejepaprovablescalableselfsupervised}
\BIBentryALTinterwordspacing
R.~Balestriero and Y.~LeCun, ``{LeJEPA}: Provable and scalable self-supervised learning without the heuristics,'' arXiv:2511.08544, 2025. [Online]. Available: \url{https://arxiv.org/abs/2511.08544}
\BIBentrySTDinterwordspacing

\bibitem{maes2026leworldmodelstableendtoendjointembedding}
\BIBentryALTinterwordspacing
L.~Maes, Q.~{Le Lidec}, D.~Scieur, Y.~LeCun, and R.~Balestriero, ``{LeWorldModel}: Stable end-to-end joint-embedding predictive architecture from pixels,'' arXiv:2603.19312, 2026. [Online]. Available: \url{https://arxiv.org/abs/2603.19312}
\BIBentrySTDinterwordspacing

\bibitem{AutomatedQualityInspection}
D.~Sopauschke, S.~Sauer, F.~Warschewske, and E.~Trostmann, ``Automated quality inspection for the manufacturing of highly customized building modules,'' in \emph{Automated Visual Inspection and Machine Vision VI}.\hskip 1em plus 0.5em minus 0.4em\relax SPIE, 2025, p. 1357202, 10.1117/12.3062391.

\bibitem{SemanticPartSegmentation}
P.~Bründl, B.~Scheffler, M.~Stoidner, H.~Nguyen, A.~Baechler, A.~Abrass, and J.~Franke, ``Semantic part segmentation of spatial features via geometric deep learning for automated control cabinet assembly,'' \emph{Journal of Intelligent Manufacturing}, vol.~35, no.~8, pp. 3681--3695, 2024, 10.1007/s10845-023-02267-1.

\bibitem{DisassemblyOfControlCabinets}
A.~Scheck, P.~Bründl, and J.~Franke, ``Deep learning-based object detection for automated disassembly of control cabinets,'' in \emph{Flexible Automation and Intelligent Manufacturing: The Future of Automation and Manufacturing: Intelligence, Agility, and Sustainability}.\hskip 1em plus 0.5em minus 0.4em\relax Cham, Switzerland: Springer Nature Switzerland, 2026, pp. 497--508, 10.1007/978-3-032-05610-8\_50.

\bibitem{DeepLearningBasedVisualControlAssistent}
M.-A. Zamora-Hernández, J.~A. Castro-Vargas, J.~Azorin-Lopez, and J.~Garcia-Rodriguez, ``Deep learning-based visual control assistant for assembly in {Industry 4.0},'' \emph{Computers in Industry}, vol. 131, p. 103485, 2021, 10.1016/j.compind.2021.103485.

\bibitem{DeepLearningAdvancedPoseEstimation}
F.~P. Basamakis, D.~Dimosthenopoulos, A.~C. Bavelos, G.~Michalos, and S.~Makris, ``Deep learning for advanced pose estimation and quality inspection in {Industry 4.0} assembly applications,'' \emph{International Journal of Computer Integrated Manufacturing}, pp. 1--20, 2026, 10.1080/0951192X.2026.2664186.

\bibitem{MonitoringAssemblyProcessUsingDeepLearning}
C.~Chen, C.~Zhang, T.~Wang, D.~Li, Y.~Guo, Z.~Zhao, and J.~Hong, ``Monitoring of assembly process using deep learning technology,'' \emph{Sensors}, vol.~20, no.~15, p. 4208, 2020, 10.3390/s20154208.

\bibitem{TowerdsGeneralizableAssemblyStateRecognition}
T.~J. Schoonbeek, G.~Balachandran, H.~Onvlee, T.~Houben, S.-H. Hung, J.~Kustra, P.~H.~N. de~With, and F.~van~der Sommen, ``Supervised representation learning towards generalizable assembly state recognition,'' \emph{IEEE Robotics and Automation Letters}, vol.~9, no.~11, pp. 9915--9922, 2024, 10.1109/LRA.2024.3468157.

\bibitem{v-jepa}
\BIBentryALTinterwordspacing
A.~Bardes, Q.~Garrido, J.~Ponce, X.~Chen, M.~Rabbat, Y.~LeCun, M.~Assran, and N.~Ballas, ``{Revisiting Feature Prediction for Learning Visual Representations from Video},'' \emph{Transactions on Machine Learning Research}, 2024. [Online]. Available: \url{https://mlanthology.org/tmlr/2024/bardes2024tmlr-revisiting/}
\BIBentrySTDinterwordspacing

\bibitem{MaskedAutoencoders}
K.~He, X.~Chen, S.~Xie, Y.~Li, P.~Dollár, and R.~Girshick, ``Masked autoencoders are scalable vision learners,'' in \emph{2022 IEEE/CVF Conference on Computer Vision and Pattern Recognition (CVPR)}, 2022, pp. 15\,979--15\,988, 10.1109/CVPR52688.2022.01553.

\bibitem{garrido2025intuitivephysicsunderstandingemerges}
\BIBentryALTinterwordspacing
Q.~Garrido, N.~Ballas, M.~Assran, A.~Bardes, L.~Najman, M.~Rabbat, E.~Dupoux, and Y.~LeCun, ``Intuitive physics understanding emerges from self-supervised pretraining on natural videos,'' arXiv:2502.11831, 2025. [Online]. Available: \url{https://arxiv.org/abs/2502.11831}
\BIBentrySTDinterwordspacing

\bibitem{ClevrChange}
D.~H. Park, T.~Darrell, and A.~Rohrbach, ``Robust change captioning,'' in \emph{2019 IEEE/CVF International Conference on Computer Vision (ICCV)}, 2019, pp. 4623--4632, 10.1109/ICCV.2019.00472.

\bibitem{zhang2026deltajepalearningactionsensitiveworld}
\BIBentryALTinterwordspacing
Z.~Zhang, Y.~Wang, Z.~Guan, Y.~Yang, B.~Shi, T.~Zong, H.~Yi, G.~Chao, X.~Chen, T.~Yang, C.~Bao, T.~Yu, J.~Zhou, and J.~Xu, ``{Delta-JEPA}: Learning action-sensitive world models via latent difference decoding,'' arXiv:2606.31232, 2026. [Online]. Available: \url{https://arxiv.org/abs/2606.31232}
\BIBentrySTDinterwordspacing

\end{thebibliography}

\end{document}